\documentclass[runningheads]{llncs}

\usepackage{eccv}

\usepackage{eccvabbrv}

\usepackage{graphicx}
\usepackage{booktabs}
\usepackage{multirow}
\usepackage{multicol}
\usepackage{makecell}
\usepackage{wrapfig}
\usepackage[normalem]{ulem}
\usepackage[accsupp]{axessibility}  
\usepackage{dsfont}

\usepackage{paralist}
\DeclareMathAlphabet      {\mathbfit}{OML}{cmm}{b}{it}

\usepackage{hyperref}

\usepackage{orcidlink}
\usepackage{makecell}

\def\sysname{GRAPE}

\usepackage{xcolor}

\begin{document}

\title{GRAPE: Generalizable and Robust Multi-view Facial Capture}

\author{Jing Li\inst{1}\and
Di Kang\inst{2} \and
Zhenyu He\inst{1} \thanks{Corresponding author: zhenyuhe@hit.edu.cn}
}

\authorrunning{J.~Li et al.}

\institute{Harbin Institute of Technology, Shenzhen \and
Tencent AI Lab\\
}

\maketitle

\begin{abstract}
Deep learning-based multi-view facial capture methods have shown impressive accuracy while being several orders of magnitude faster than a traditional mesh registration pipeline.
However, the existing systems (e.g. TEMPEH) are strictly restricted to inference on the data captured by the same camera array used to capture their training data.
In this study, we aim to improve the generalization ability so that a trained model can be readily used for inference (i.e. capture new data) on a different camera array.
To this end, we propose a more generalizable initialization module to extract the camera array-agnostic 3D feature, including a visual hull-based head localization and a visibility-aware 3D feature aggregation module enabled by the visual hull.
In addition, we propose an ``update-by-disagreement'' learning strategy to better handle data noise (e.g. inaccurate registration, scan noise) by discarding potentially inaccurate supervision signals during training.
The resultant \underline{\textbf{g}}eneralizable and \underline{\textbf{r}}obust topologically consistent multi-view facial c\underline{\textbf{ap}}tur\underline{\textbf{e}} system (\sysname{}) can be readily used to capture data on a different camera array, reducing great effort on data collection and processing.
Experiments on the FaMoS and FaceScape datasets demonstrate the effectiveness of the proposed method.
\end{abstract}

\section{Introduction}
\label{sec:intro}

Reconstructing a high-quality 3D human face geometry from synchronized multi-view images has been studied for years due to the great demand in the film and game industries.

Conventional workflows in this domain typically consist of two steps: first, acquiring raw scans using multi-view stereo (MVS) techniques, and then wrapping a template mesh (in a predefined topology) to the scan.
There exist two major drawbacks in the conventional data processing pipeline: 1) MVS requires a dense camera array setup and is usually slow; 2) a huge amount of manual work is required, including cleaning the scan noise, completing the partial scans, selecting facial key points for registration, etc.
Recently, there have been a few attempts to simplify and speed up the process by directly predicting the registered mesh~\cite{tofu,tempeh}, demonstrating comparable accuracy while being significantly more efficient.

\par
ToFu~\cite{tofu} is the first work to utilize a data-driven method to directly predict a fixed topology mesh from calibrated multi-view images.
However, the data used to train ToFu~\cite{tofu} is obtained using the conventional workflow.
Later, TEMPEH~\cite{tempeh} proposes to only use \emph{automatically} registered meshes (i.e. FLAME tracking), resulting in greatly reduced data processing time and cost.
To remedy potential errors in terms of shape similarity and vertex correspondences, TEMPEH introduces two major changes: 
1) uses an additional point-to-surface distance between the raw scan points and the predicted FLAME mesh (since the tracked FLAME may not be accurate); 
2) trains their model on much larger-scale dynamic data (since it is easier to maintain consistent correspondence).
However, the trained network can only handle data captured using the same camera setup as the training data, which means we need to re-collect data (e.g. 95 subjects, 28 expressions for each subject, 600K frames in total~\cite{tempeh}) and perform mesh tracking if we have a different camera array.
\par
In this work, we aim to improve its generalization so that a new camera array can be quickly used to conduct fast topologically consistent facial performance capture without the expensive and time-consuming data collection and processing as in TEMPEH.
After training on one dataset (e.g. FaMoS~\cite{tempeh}), our network achieves a good reconstruction accuracy when directly applied to another dataset (e.g. FaceScape~\cite{facescape}) for inference.
The accuracy is further improved after finetuned with a small amount of data from the target camera array.

The fundamental reason that prevents TEMPEH~\cite{tempeh} from generalizing to new camera array data is that TEMPEH is a completely data-driven system and overfits the training condition.
We find the key to better generalization is to obtain a camera array-agnostic 3D feature cube.
To this end, we first replace its learning-based head localizer with a visual hull.
The visual hull is calculated from the multiview head masks through unprojection and is thus applicable for different camera arrays.
The coarse shape depicted by the visual hull not only provides a good head localization but also enables the following visibility-aware 3D feature aggregation.
This visibility is another key factor in obtaining a more generalizable 3D feature cube.
In TEMPEH's global network, a voxel feature is obtained by aggregating 2D image features from both the frontal images and the back images, introducing substantial noise.
This is not a severe problem when training and testing the model on data captured by a fixed camera array as TEMPEH.
However, it prevents the trained model from generalizing to a different camera array (e.g. FaceScape).
In contrast, our visibility-aware 3D feature aggregation largely alleviates this issue, resulting in better generalization.

\par
Another practical issue is to properly handle the training data noise, including incomplete scans, scan noise at the back of the head, and registration errors caused by the fully automatic mesh registration.
To this end, we propose a novel ``update-by-disagreement'' training strategy inspired by \cite{coteaching,coteaching_plus,decoupling}.
Specifically, we design a loss to automatically discard potential incorrect supervision signals by comparing the prediction from two networks and the supervisory scan/mesh.
The supervision is considered inaccurate and discarded if the two predictions are close to each other and are distant from the supervision, which means the networks are updated if their predictions are different enough (``disagreement'').

In summary, our contributions include:
\begin{compactitem}
    \item We propose the first topologically consistent multi-view head capture method that can easily generalize to new camera arrays and can robustly train on data with registration error or scan noise.
    \item We propose to utilize visual hull and visibility-aware 3D feature aggregation to replace the learned head localizer in TEMPEH, which is the key to a generalizable method.
    \item We propose an ``update-by-disagreement'' learning strategy to automatically discard potentially noisy supervision, effectively improving the training robustness.
    \item Experiments show that our method works well on noisy data and could get acceptable reconstruct accuracy when applying the trained model on a different camera array.
\end{compactitem}

\section{Related Work}
\label{sec:related_work}
\noindent\textbf{Multiview face capture.}
Reconstructing 3D human faces from multi-view systems is a long-standing problem and has been studied for many years.
Most methods require two steps to get a registered human face, that is face shape acquisition and face registration~\cite{3dmm_review,1467528,1240231}.
Early multiview-based methods get the face shape by matching image features from different views, which is time-consuming and tends to fail if the camera array is sparse.
In recent years, many deep learning-based 3D MVS methods~\cite{mvsnet,transmvsnet,kdmvs} have shown great success in the sparse view scenario.
However, the above methods only predict the unstructured scans.
Mesh register algorithms are necessary for the above methods for a face capture system, which would introduce registration error to the final topology consistent mesh.
Recently, several real-time learning-based face capture systems have been proposed to directly predict the registered mesh from the multiview images~\cite{tofu,tempeh}. 
However, the generalization and the robustness of the learning-based methods~\cite{tofu,tempeh} are not as good as the geometric-based methods~\cite{3dmm_review}.
Our method alleviates the problem by introducing a geometric-based initialization.
\par
\noindent\textbf{Monocular face reconstruct.}
There has been a surge of recent interest in reconstructing the human head from monocular image~\cite{flame,ringnet,deca,3dmm,expnet,mofa,Genova,inverse_rendering,face2face}.
Most methods solve the ill-posed problem in an analysis-by-synthesis fashion, which reconstructs the human face by estimating the parameters of a statistical model (e.g. 3DMM~\cite{3dmm}) to minimize the difference between the observed images and the synthesized images.
In the early years, most methods predicted the parameters in optimizing based method~\cite{inverse_rendering,3dmm,face2face}.
Recently, most deep learning-based methods directly predict the parameters from the input image~\cite{deca,expnet,Genova,ringnet,mofa}.
However, the reconstruction accuracy of the algorithms is limited by the expression ability of the statistical model.
\par
Some model-free methods attempt to directly predict faces from images~\cite{vrn,e2far,unsup3d}.
However, their performance is still limited by the ambiguity of the monocular image.
What's more, these methods often predict the unstructured meshes, which still require to be registered to a common topology.
Instead, our method also directly predicts the meshes with a high geometry accuracy and in a common topology.
\par
\noindent\textbf{Learning from noise.}
Due to the noise data being widespread and difficult to clean, there has been a surge of attempts to learn models from noisy data.
In general, these methods can be divided into three categories, respectively, which are based on regularization~\cite{manifold_regular,VirtualAT}, noise transition matrix~\cite{masking}, and sample selection~\cite{mentornet,coteaching,decoupling,coteaching_plus}.
Regularization-based methods often can not get the optimal performance~\cite{masking}.
Noise transition matrix-based methods are hard to adapt to our task.
\par
Sample selection-based methods assume that the network may give a correct prediction for data with noisy label~\cite{mentornet,coteaching,coteaching_plus}.
Co-teaching~\cite{coteaching} assumes that samples with small losses have a lower probability of belonging to noise, and proposes a small-loss trick.

They train two networks at the same time and the parameters of each network are updated with the small loss data of the other network.
Decoupling~\cite{decoupling} proposes an ``update-by-disagreement'' paradigm, where parameters are updated only when two networks can not reach a consensus.
Co-teaching+~\cite{coteaching_plus} shows that ``update-by-disagreement'' can be combined with ``small loss trick'' to achieve state-of-the-art results.
All the above methods are designed for classification tasks.
Our method adapts the core idea of ``small loss trick'' and ``update-by-disagreement'' to face reconstruction tasks.

\section{Method}
\begin{figure}[t]
\centering
\includegraphics[width=\textwidth]{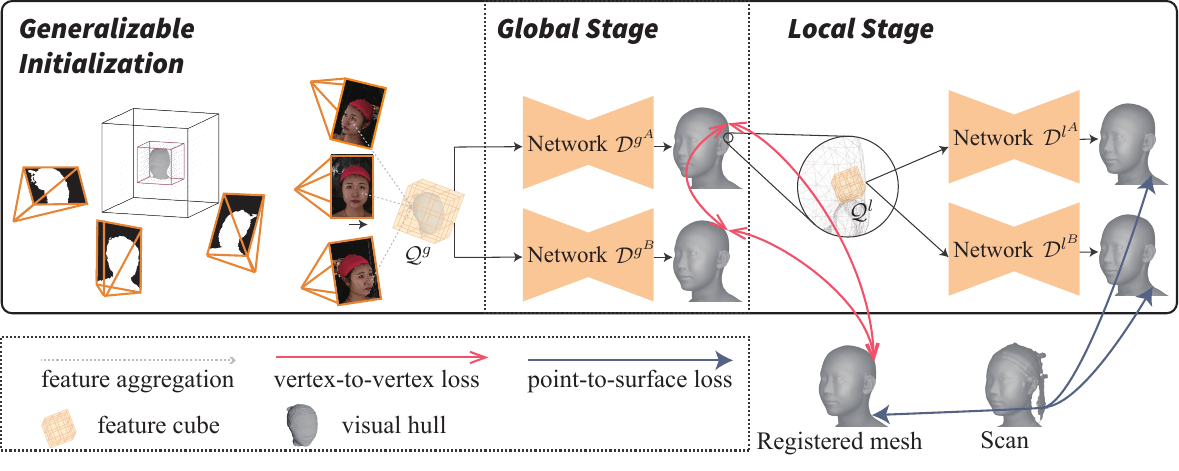}
\caption{
\textbf{Overview of \sysname{}.} 
\sysname{} consists of a non-data-driven and thus generalizable initialization module, and two 3D ConvNets learned in a coarse-to-fine manner to predict the mesh vertex locations.
(1) The initialization module first calculates a visual hull from the foreground masks
and then aggregates 2D image features with the consideration of visibility, resulting in a camera setup-agnostic 3D feature cube generalizable to different camera arrays.
(2) The networks are trained with an ``update-by-disagreement'' learning strategy to handle registration errors and scan noises, where two networks are used in each stage to discard unreliable supervision signals.
}
\label{fig:overview}
\end{figure}

Given a set of synchronized images $\mathcal{I}=\{I_1, I_2, ..., I_N\}$ captured by a camera array and their corresponding camera parameters $\mathcal{P}=\{P_1, P_2, ..., P_N\}$, \sysname{} (Fig.~\ref{fig:overview}) predict per-vertex locations of a mesh $M$ in a predefined topology.

As shown in Fig.~\ref{fig:overview}, \sysname{} consists of three major steps.
First, a non-data-driven initialization method to produce camera array-agnostic 3D feature cube $\mathbfit{Q}^g$, including projection-based visual hull calculation and visibility-aware 3D feature aggregation (Sec.~\ref{sec:vishull}).
The feature cube $\mathbfit{Q}^g$ is then processed by a 3D ConvNet $\mathcal{D}^g$ (Sec.~\ref{sec:global}) to obtain the vertex positions of the mesh in a predefined topology (mesh $\mathcal{M}^g$).
Finally, another 3D ConvNet $\mathcal{D}^l$ refines the per-vertex position of $\mathcal{M}^g$ with only the zoomed-in surrounding features $\mathbfit{Q}^l$ of that vertex (Sec.~\ref{sec:local}). 
To robustly handle commonly seen data noise,
both the networks are trained with the proposed ``update-by-disagreement'' strategy, which requires two networks to test their \emph{agreement} (Sec.~\ref{sec:learning}).
\label{sec:method}

\subsection{Generalizable initialization}
\label{sec:vishull}
It is very beneficial to properly initialize a 3D feature cube for later network training
(analogous to using a center-cropped face image for landmark detection).
We propose to localize the head with a pure projection-based visual hull algorithm agnostic to different camera setups.
Compared to using a learned localizer~\cite{tempeh} to produce an oriented bounding box, using the visual hull as initialization is not only suitable for different camera setups but also enables us to adopt visibility-aware 3D feature cube aggregation on the global stage.
These two operations constitute the most crucial module that makes \sysname{} successfully generalize to different camera setups.
\subsubsection{Visual hull for localization.}
Our method localizes the head position from a visual hull.
Specifically, a discrete voxel grid $\mathbfit{G}^h\in \mathbb{R}^{d^h\times d^h\times d^h\times 3}$ storing zeros is initialized and projected onto every camera view using the calibrated camera parameters $\mathcal{P}$.
A voxel is considered seen by an image view $I_i$ if and only if its projected location belongs to the foreground mask of this view $I_i$.
We conduct such projection tests for every view and count how many times every voxel has been seen.
Finally, a marching cube algorithm is applied on the occupancy grid (storing counts) and produces a visual hull $\mathcal{H}$ of the object.
The visual hull can reliably provide us with a head bounding box.

\subsubsection{Visibility-aware 3D volume feature aggregation.}

\begin{figure}[t]
\centering
\begin{subfigure}{0.4\textwidth}
\centering
\includegraphics[width=\textwidth]{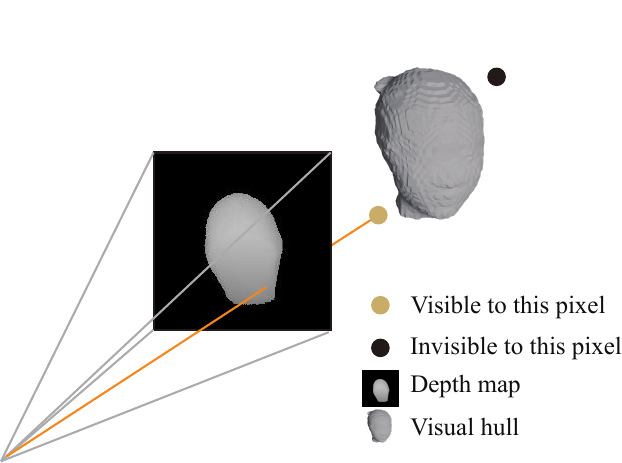}
\caption{Visibility check for one voxel.}
\label{fig:voxel_visibility}
\end{subfigure}
\begin{subfigure}{0.58\textwidth}
\centering
    \includegraphics[width=0.6\textwidth]{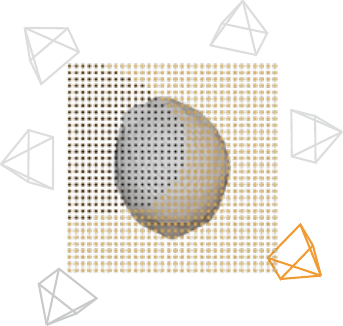}
    \caption{Visibility of the provided view to all voxels.}
    \label{fig:voxel_grid_visibility}
\end{subfigure}
\label{fig:visibility}
\caption{
\textbf{Per-voxel visibility check.} 
The provided input view is visible for all the yellow voxels.
A voxel aggregates image features only from visible views by aggregating 2D image features from its projected location.
}
\end{figure}

Visibility plays a crucial role when handling data captured by different camera arrays since using 3D features polluted by images from invisible views (e.g. from the opposite side) negatively affects the accuracy even for a fixed camera setup (Tab.~\ref{tab:sota_famos}\&\ref{tab:sota_facescape}).
Similar to TEMPEH, we calculate per voxel mean and variance features for the \emph{visible} views only.
As shown in Fig.~\ref{fig:voxel_visibility}, the visibility of an image view $I_i$ to a voxel $o_{j,k,l}\in \mathbfit{G}^g$ is determined by comparing the voxel's depth to the optical center (\emph{voxel depth} for brevity) $d_i^{o_{j,k,l}}$ and the depth value $d_i^{\mathcal{H}(r(o_{j,k,l}))}$ of its projected pixel location on the depth map rendered with the mesh $\mathcal{H}$ (\emph{mesh depth} for brevity). 
A voxel $o_{j,k,l}$ is visible to view $I_i$ if and only if its voxel depth is smaller than its corresponding mesh depth. 
The visibility of all the voxels (2D illustration) for camera $I_i$ is shown in Fig.~\ref{fig:voxel_grid_visibility}.
To account for the imprecision of the estimated visual hull $\mathcal{H}$, 
we include a margin in the comparison as follows,
{
\scriptsize
\begin{align}
    \mathds{1}_{i,j,k,l}=\left\{
    \begin{aligned}
        0, \Vert d_i^{o_{j,k,l}} - d_i^{\mathcal{H}(r(o_{j,k,l}))} \Vert > \rho \\
        1,  \Vert d_i^{o_{j,k,l}} - d_i^{\mathcal{H}(r(o_{j,k,l}))} \Vert \leq \rho
    \end{aligned}
\right.
\end{align}
}
where $d_i^{o_{j,k,l}}$ is the depth value of voxel $o_{j,k,l}$ from view $I_i$,
$d_i^{\mathcal{H}(r(o_{j,k,l}))}$ is the depth of the ray $r(\cdot)$ starting from the optical center and passing through the voxel $o_{j,k,l}$,
$\rho$ is the margin that controls the visibility of the voxels.

Similar to \cite{tempeh}, a voxel's feature is the concatenation of the mean and variance of its corresponding 2D image features sampled from all the visible views.
Specifically, we use a voxel grid with predefined resolution $\mathbfit{G}^g\in \mathbb{R}^{d^g\times d^g\times d^g \times3 }$ and put it at the geometrical center of the visual hull $H$.
A voxel $o_{j,k,l}\in \mathbfit{G}^g$ is projected to view $I_i$ using the calibrated camera parameters $P_i$ to extract 2D image feature $f_{i,j,k,l}$ with bilinear sampling from the feature map $F_i$.
Formally $f_{i,j,k,l}=\theta(F_i,\Pi(o_{j,k,l}, P_i))$, where $\theta(\cdot)$ is the bilinear sampling operation, $\Pi(\cdot)$ the projection.
The feature vectors of different views are then fused by calculating the weighted mean and variance:
{
\scriptsize
\begin{align}
    \mu_{j,k,l}=&\frac{1}{\sum_{i=1}^{n^v} \mathds{1}_{i,j,k,l}}\sum_{i=1}^{n^v} (\mathds{1}_{i,j,k,l}\times f_{i,j,k,l}), \\
    \sigma_{j,k,l}^2=&\frac{1}{\sum_{i=1}^{n^v} \mathds{1}_{i,j,k,l}}\sum_{i=1}^{n^v} (\mathds{1}_{i,j,k,l}\times f_{i,j,k,l})^2-(\mu_{j,k,l})^2
\end{align}
}
where $n^v$ is the number of views.

\par
\subsection{Vertex prediction}
\label{sec:network}

Following the coarse-to-fine strategy~\cite{tempeh}, we also use a global stage network, which takes as input the complete 3D feature cube $Q^g$ as input and predicts per-voxel probability for all the mesh vertices, and a local stage network, which refines a vertex using only the local 3D volume feature surrounding it.
The global and local stages use 3D UNet-style networks as in \cite{tempeh}.

\subsubsection{Global stage.}
\label{sec:global}
In the global stage, the network $\mathcal{D}^g$ predicts a coarse mesh $\mathcal{M}^c$ from the complete 3D feature cube $\mathbfit{Q}^g$.
Specifically, the 3D feature cube $\mathbfit{Q}^g$ is first transformed into a cost volume $\mathbfit{C}\in \mathbb{R}^{n^v\times d^g\times d^g \times d^g}$, which is followed by a softmax activation (across all spatial locations) to predict the per-vertex probability for all the mesh vertices.
The final location for a vertex $V_i$ is the weighted sum of the voxel location and its voxel probability.

\subsubsection{Local stage.}
\label{sec:local}
In the local stage, the network $\mathcal{D}^l$ refines a vertex's location using the local 3D feature cube surrounding it as in TEMPEH~\cite{tempeh}.
Specifically, we use a smaller local voxel grid $\mathbfit{G}_i^l \in \mathbb{R}^{d^l\times d^l \times d^l \times 3}$ centered at vertex $v_i$ to extract local 3D features from the 2D image feature maps.
The edge length of a voxel is 2mm.
Different from the 3D feature aggregation operation, the visibility is tested only once per view for the vertex $v_i$ and used for all the voxels in the local grid.
In addition to visibility, the cosine value of the angle between the vertex normal and the view direction is used as a weight during the feature aggregation of $\mathbfit{Q}^l$.
Then, the per-voxel probability of the current voxel $v_i$ is predicted by the local network $\mathcal{D}^l$ using $\mathbfit{Q}^l$.

\subsection{Training with noisy data}
\label{sec:learning}

It is common for the scans to have noise in the back of the head.
Thus, it is necessary to explicitly handle such noise when combining multiple datasets for training.
Inspired by ~\cite{decoupling,coteaching,coteaching_plus}, we propose a ``update-by-disagreement'' learning strategy, resulting in a \emph{robust vertex-to-vertex distances} to handle registration errors and a \emph{robust point-to-surface distance} to handle scanning noise.
Both the robust vertex-to-vertex distance $E_\text{v2v}$ and the robust point-to-surface distance $E_\text{p2s}$ are applied during the coarse stage training.
At the refining stage, \sysname{} mainly focuses on predicting an accurate shape, thus only the robust point-to-surface distance $E_\text{p2s}$ is applied.

To utilize the proposed ``update-by-disagreement'' learning strategy, we use two networks (same architecture, different weight initialization) in both the global and the local stages and update the network parameters based on the similarity of their predictions.

\par
\subsubsection{Robust vertex-to-vertex supervision from the registered meshes.}
\label{sec:reg}
It is required to use registered meshes for training, where the meshes are in the same topology and the vertices with the same index on different meshes maintain the same semantic meaning (e.g. nose tip, left eye corner) across different shapes.
However, registration error broadly exists in the existing datasets, especially for automatically registered datasets (e.g. FaceScape).
To alleviate this issue, we propose a novel robust vertex-to-vertex distance that automatically ignores the vertices that may have low registration qualities before back-propagating their gradients.
Specifically, the parameter of network $\mathcal{D}^{g^A}$ is updated with the following loss:
{
\scriptsize
\begin{align}
    E_\text{v2v}^\text{A}&=\frac{1}{\sum_{i=1}^{n^v} \delta_i} \sum_{i=1}^{n^v} \delta_i \Vert v_i^\text{A}-v^\text{reg}_i \Vert,\\
    \delta_i&=\left\{
    \begin{aligned}
        0, \Vert v_i^\text{B} -v_i^\text{A}  \Vert < \Vert v_i^\text{B} -v_i^\text{reg} \Vert \\
        1, \Vert v_i^\text{B} -v_i^\text{A}  \Vert \geq \Vert v_i^\text{B} -v_i^\text{reg} \Vert 
    \end{aligned} \nonumber
\right.
\end{align}
}
where $n^v$ is the vertex number, $v_i$ is the position of $i$th vertex.
Vertices predicted from networks $\mathcal{D}^{g^\text{A}}$ and $\mathcal{D}^{g^\text{B}}$ are noted as  $v^\text{A}$ and $v^\text{B}$, respectively. $v^\text{reg}$ is the vertices of the registered mesh.
$\delta_i$ indicates whether the two networks make a consensus estimation for the $i$th vertex.
With this update-by-disagreement filtering strategy, the gradient from $v_i^\text{reg}$ will be discarded if the vertex locations predicted by both the networks are close to each other while being far away from the supervision $v_i^\text{reg}$, which is potentially noisy supervision.
Network $\mathcal{D}^{g^\text{B}}$ is updated similarly and its description is omitted for clarity.

\par
In the beginning, almost all the predicted vertices are updated according to the registered mesh since the two randomly initialized networks produce very different results.
As the training continues, the two networks gradually converge into similar states and their predictions become similar.
Since it is easier to learn the correct labels than the noisy labels, this update-by-disagreement strategy can better prevent the network from being negatively affected by the noisy labels, resulting in more robust training on noisy training data.

\subsubsection{Robust point-to-surface supervision from the scan.}
Using the original scan as supervision can potentially improve the shape accuracy 
since the automatically registered FLAME meshes usually contain errors due to the limited expressiveness of the learned 3DMM prior and limited vertices. 
However, raw scans usually contain noise, especially in the back of the head.
We propose to use a similar ``update-by-disagreement'' training strategy on the point-to-surface distance.

Point-to-surface distance for a given point $p_i$ of scan $\mathcal{S}$ to a mesh $\mathcal{M}$ is defined as the minimum distance of $p_i$ to the surface of $\mathcal{M}$.
Mathematically, the distance of $i-$th point of the scan to the mesh is calculated as 
{
\scriptsize
\begin{equation}
    d_i^\mathcal{M}=\mathop{\min}_{m\in \mathcal{M}} \Vert p_i-m \Vert.
\end{equation}
}
The \emph{robust} point-to-surface distance to update network $\phi_A$ is calculated by:
{
\scriptsize
\begin{align}
    E_\text{p2s}^\text{A}&=\frac{1}{\sum_{i=1}^{n^p}\omega_i}\sum_{i=1}^{n^p} \omega_i d_i^A\\
    \omega_i&=\left\{
    \begin{aligned}
        & 0,\;\; \left(d_i^\text{B} < d_i^\text{reg}\right)  \cap 
 \left(\left|d_i^\text{B} - d_i^\text{A} \right| < \left| d_i^\text{B} - d_i^\text{reg} \right|\right)   \\
        & 1,\;\; \text{otherwise} 
    \end{aligned} \nonumber
\right.
\end{align}
}
where $n^p$ is the point number of the scans, 
$d_i^\text{A}$ and $d_i^\text{B}$ the point-to-surface distance between  $p_i$ and the predicted mesh $\mathcal{M}^A$ ($\mathcal{M}^B$) from network $\mathcal{D}^A$ ($\mathcal{D}^B$),
$d_i^\text{reg}$ is distance between $p_i$ and the registration mesh $\mathcal{M}^{reg}$.

\par
Heuristically, point $p_i$ is probably an outlier, which will not be used for network update in this iteration if it is closer to the predicted mesh $\mathcal{M}^B$ than the registered mesh $\mathcal{M}^{reg}$, and the predicted meshes $\mathcal{M}^A$ and $\mathcal{M}^B$ are close to each other.

\begin{figure}[t]
\centering
\includegraphics[width=\textwidth]{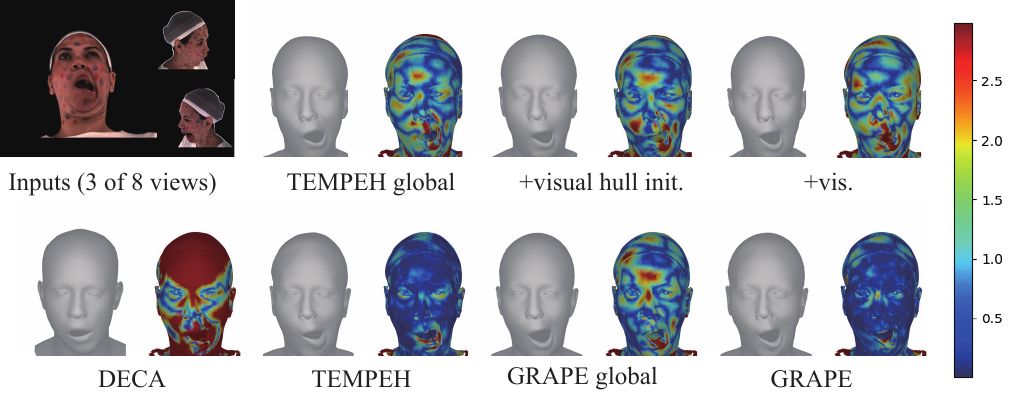}
\includegraphics[width=\textwidth]{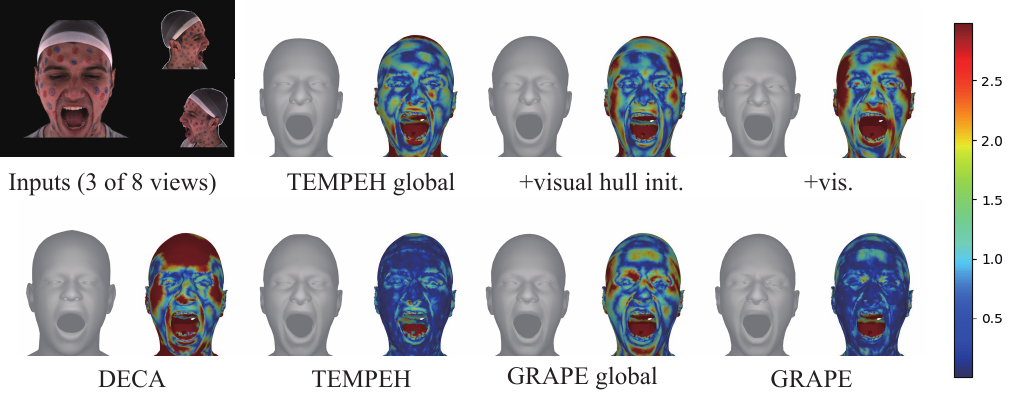}
\caption{\textbf{Qualitative comparisons on the FaMoS dataset.} 
For each model, we show the predicted mesh (left) and the raw scan mesh (right) colored with the point-to-surface distance. The corresponding color bar (3 mm max.) is shown on the right.
}
\label{fig:famos_geometry_accuracy}
\end{figure}
\begin{figure}[t]
\centering
\includegraphics[width=\textwidth]{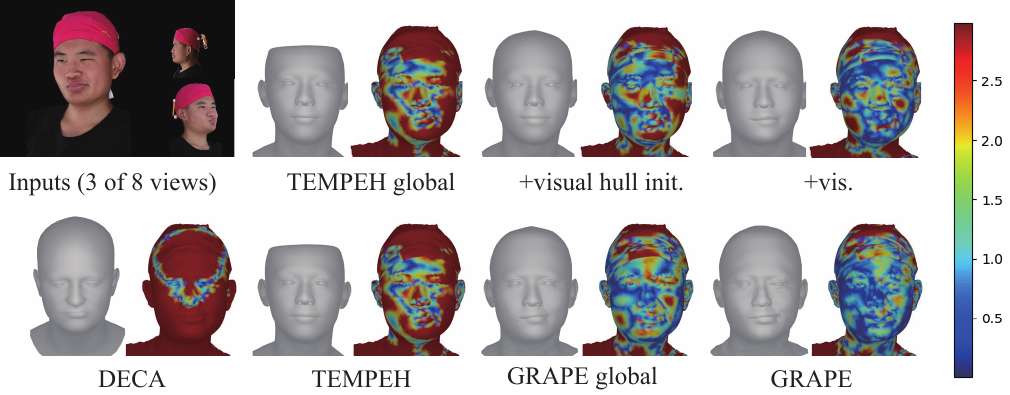}
\includegraphics[width=\textwidth]{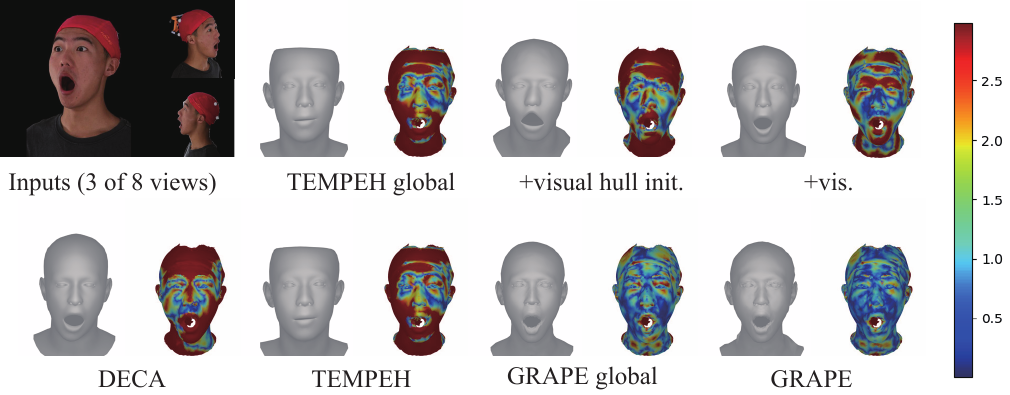}
\caption{\textbf{Qualitative comparisons on the FaceScape dataset.}
For each model, we show the predicted mesh (left) and the raw scan mesh (right) colored with the point-to-surface distance. The corresponding color bar (3 mm max.) is shown on the right.
}
\label{fig:fs_geometry_accuracy}
\end{figure}

\section{Experiment}
\label{sec:exp}
\noindent\textbf{Datasets.}
Our experiments are conducted on FaMoS~\cite{tempeh} and FaceScape~\cite{facescape} datasets.
FaMoS dataset records 95 subjects, each performing 28 predefined actions (e.g. anger, disgust, fear, surprise).
They have provided registered FLAME~\cite{flame} meshes with high accuracy.
Following TEMPEH~\cite{tempeh}, 88478/1118/8350 frames are sampled from 70/8/15 subjects for training/validation/test. 
\par
The FaceScape dataset captures 20 different expressions of 938 subjects captured via a dense multi-view system that includes 68 DSLR cameras.
We use the publicly available subset of the dataset, which includes 359 subjects, each containing 20 different expressions\footnote{Several expressions are missing in the release data.}.
Note that FaceScape captures only one static frame per expression not a sequence of data~\cite{tempeh}.
One subject's data has been held out for the experiments that require fine-tuning.
The remaining data are randomly split into train/test sets,
resulting in 6960 expressions (from 348 subjects) for training and 218 expressions (from 11 subjects) for testing.
We conduct FLAME fitting~\cite{flame} on their original registered meshes (in a different topology).
We only report face region accuracy on FaceScape because the scalp and neck regions of the scans in FaceScape are too noisy (even incomplete) and cannot be used for evaluation.

\par
\noindent\textbf{Implementation details.}
Following TEMPEH~\cite{tempeh}, we use a UNet~\cite{unet} feature extractor with a ResNet34~\cite{resnet} backbone.
To improve generalization,
we only update the 4 transposed convolution layers used for upsampling while keeping the ResNet backbone fixed.
%
The masks are obtained with PP-Matting~\cite{ppmatting} and filtered by only keeping the largest connected component.
The same data processing is applied during inference to get a proper visual hull as initialization.
The segmentation results are further cleaned by only keeping the largest connected region.
All the images are padded to squares and then resized to $600\times600$.
For the FaMoS dataset, we take all 8 color images as input.
For the FaceScape dataset, we randomly select 8 images during training.
Data augmentation including color jitters, random rotation, and random scale are applied during training.
The grid used for visual hull calculation is at $160^3$ resolution and its voxel edge length is set to 5 mm, resulting in a similar total size to the grid estimated by the localizer TEMPEH.
The size of the feature grid of the global stage is $32^3$, and the local stage is $8^3$.
$\rho$ in Eq. 1 is set as 0.1.
The 3D ConvNet used to process the feature volume cube 
shares the same architecture with TEMPEH~\cite{tempeh}.
When fine-tuning data from another camera setup, we only update the 4 transposed convolution layers used to upsample the ResNet feature.
AdamW optimizer~\cite{adamw} is adopted in our experiments, with a learning rate of 0.0001.
The global stage takes 12G memory (batch size=2) and the local stage takes 22G memory (batch size=1).
Our method is implemented with PyTorch~\cite{pytorch}.

\par
\noindent\textbf{Baselines.}
TEMPEH~\cite{tempeh} is a state-of-the-art multi-view topology consistency head capture algorithm.
DECA~\cite{deca} is a single-image face reconstruction algorithm that directly predicts the FLAME~\cite{flame} parameter. 
Note that we select the most accurate result (with Procrustes Analysis and scale) predicted from all input views to calculate its performance, serving as another reference.
We also introduce two variations of the TEMPEH model as the baseline to analyze the influence of the proposed modules, denoted as ``+ V.H. init.'' and ``+vis.''.
``+V.H. init. '' replaces the learning-based localizer in TEMPEH with the bounding box derived from the visual hull. 
``+vis.'' model \emph{further} includes the visibility-aware 3D feature aggregation.
Note that we fix the ResNet layers and only tune the additional upsampling convolution layers (same as \sysname{}) in the 2D image feature extractor for these two ablation studies.
Further equipping ``+vis.'' with the proposed ``update-by-disagreement'' training strategy obtains our \sysname{}.

\noindent\textbf{Evaluation metric.} 
Following TEMPEH~\cite{tempeh}, we evaluate the shape accuracy with point-to-surface distance (from the GT scan to the predicted FLAME mesh).
On FaMoS~\cite{tempeh}, the scans are segmented into different regions to analyze the accuracy of each part.
On FaceScape~\cite{facescape}, we only evaluate the accuracy of the face region due to scan noise.

\begin{table}[t]
\scriptsize
\caption{\textbf{Result on FaMoS dataset.} 
We evaluate point-to-scan errors on the FaMoS dataset. 
DECA is not trained on the dataset like other methods. Errors in mm.}
\label{tab:sota_famos}
\centering
\begin{tabular}{l|l|rrr|rrr|rrr|rrr}
\toprule
       & & \multicolumn{3}{c|}{Complete Head} &  \multicolumn{3}{c|}{Face} &  \multicolumn{3}{c|}{Scalp} &  \multicolumn{3}{c}{Neck} \\
     &  Method & Med. & Avg. & Std. & Med. & Avg. & Std. & Med. & Avg. & Std. & Med. & Avg. & Std. \\
\midrule
    a.1 & TEMPEH~\cite{tempeh} global & 1.93 & 2.11 & \textbf{0.39} & \textbf{1.15} & \textbf{1.19} & 0.39 & 2.21 & 2.29 & 0.96 & \textbf{2.40} & 2.75 & 1.45 \\
    a.2& + V.H. init. & 2.08 & 2.38 & 0.87 & 1.25 & 1.40& 0.50 & 2.05 & 2.33 & 0.95 & 2.68 & 3.56 & 2.61 \\
    a.3 & \hspace{1mm} + vis. & 1.92 & 2.05 & 0.57 & 1.18 & 1.26 & 0.42 & 1.72 & 1.99 & 0.87 & 2.58 & 2.98 & 1.59 \\
      a.4& \hspace{2mm} + UBD (\sysname{} global) & \textbf{1.68} & \textbf{1.72} & \textbf{0.39} & \textbf{1.15} & 1.20 & \textbf{0.22} & \textbf{1.21} & \textbf{1.36} & \textbf{0.42} & 2.59 & \textbf{2.50} & \textbf{0.98} \\
      \midrule
      b.1&  TEMPEH~\cite{tempeh} (local) & 0.79 & 0.85 & 0.32 & \textbf{0.39} & \textbf{0.43} & 0.30 & \textbf{0.59} & \textbf{0.63} & \textbf{0.29} & 1.22 & 1.39 & 0.79 \\
      b.2& \sysname{} (local) & \textbf{0.75} & \textbf{0.81} & \textbf{0.20} & 0.42 & 0.44 & \textbf{0.09} &  0.67 & 0.76 & 0.30 & \textbf{1.00} & \textbf{1.04} & \textbf{0.31} \\
      \midrule
      \midrule
      c.1 &  Registration & 0.54 & 0.77 & 0.53 & 0.12 & 0.15 & 0.28 & 0.33 & 0.84 & 0.90 & 0.62 & 1.16 & 1.24\\
      c.2 & DECA~\cite{deca} & 6.96 & 7.25 & 2.27 & 2.78 & 3.33 & 1.79  & 6.41 & 6.50& 2.11 & 13.51& 13.36 & 5.47\\
\bottomrule
\end{tabular}
\end{table}
\begin{table}[t]
\scriptsize
\caption{\textbf{Results on FaceScape dataset.} 
We evaluate point-to-scan errors on the face region since the provided scans in the FaceScape dataset contain obvious noise and are sometimes incomplete on the scalp and neck region.
DECA is not trained on the dataset like other methods. Errors in mm.}
\label{tab:sota_facescape}
\centering
\setlength{\tabcolsep}{18pt}{
\begin{tabular}{l|l|ccrrr}
\toprule
   &   Method &  Med. & Avg. & Std. \\
        \midrule
a.1 &      TEMPEH~\cite{tempeh} global & 2.91 & 3.10 & 0.95 \\
a.2 &         + V.H. init. & 2.29 & 2.53 & 1.09 \\
a.3 &        \hspace{1mm} + vis. & 2.25 & 2.54 & 1.08 \\
a.4 &      \hspace{2mm} + UBD (\sysname{} global) & \textbf{1.05} & \textbf{1.20} & \textbf{0.55} \\
        \midrule
b.1 &      TEMPEH~\cite{tempeh} & 2.88 & 3.10 & 0.95 \\
b.2 &     \sysname{} & \textbf{0.76} & \textbf{0.80} & \textbf{0.18} \\
     \midrule\midrule
c.1 &      Registration & 1.29 & 1.37 & 0.61 \\
c.2 &      DECA~\cite{deca} &  5.88 & 5.98 & 1.85\\
\bottomrule
\end{tabular}
}
\end{table}
\begin{table}[t]
\centering
\scriptsize
\caption{Results on FaceScape dataset with different margin settings.}
\setlength{\tabcolsep}{30pt}{
\begin{tabular}{c|ccc}
\toprule
     Margin $\rho$ &Med.$\downarrow$ & Avg. $\downarrow$ & Std. $\downarrow$\\
     \midrule
     0.1 & \textbf{1.05} & \textbf{1.20} & 0.55\\
     \midrule
     0.05 & 1.07 & 1.25 & 0.65 \\
     0 & 1.08 & 1.23 & \textbf{0.45} \\
     \bottomrule
\end{tabular}
}
\label{tab:rho}
\end{table}

\begin{table}[t]
\scriptsize
\caption{\textbf{Model generalization.} The models are trained on the FaMoS (FaceScape) dataset and evaluated on the FaceScape (FaMoS) dataset.
Face region errors (in mm) are reported.
}
\label{tab:generalization}
\centering
\setlength{\tabcolsep}{2.5pt}{
\begin{tabular}{l|l|rrr|rrr|rrr}
\toprule
&  & \multicolumn{3}{c|}{ \makecell{w/o \\finetuning} } &  \multicolumn{3}{c|}{\makecell{finetuned \\on Subj. A}} &  \multicolumn{3}{c}{ \makecell{finetuned\\ on Subj. B}} \\
\midrule
& Method & Med. & Avg. & Std. & Med. & Avg. & Std. & Med. & Avg. & Std. \\
\midrule

\multirow{5}{*}{ \makecell{FaMoS $\rightarrow$ FS}} &TEMPEH global & 32.49& 32.27 & 13.47 & 3.61 & 7.82 & 11.84 & 3.07 & 3.92 & 3.48\\
 &+ V.H. init. & 5.77 & 8.06 & 6.65 & 3.12 & 3.96 & 2.81 & 3.68 & 4.32 & 3.22 \\
 & \hspace{1mm} + vis. & 5.95 & 7.20 & 4.44& 2.36 & 2.90 & 2.05 & 2.14 & 2.62 & 1.56 \\
     & \hspace{2mm} + UBD (\sysname{} global) & \textbf{1.92} & \textbf{2.43} & \textbf{2.86} & \textbf{2.14} & \textbf{2.37} & \textbf{0.87} & \textbf{2.16} & \textbf{2.28} & \textbf{0.75}\\
\midrule
\multirow{4}{*}{\makecell{FS $\rightarrow$ FaMoS}}& TEMPEH global & 25.63 & 32.26 & 21.77 &  5.17 & 6.56 & 4.85 & 5.44 & 5.87 & 3.30 \\
& + V.H. init. & 5.61 & 6.06 & 2.64 & 5.92 & 7.75 & 4.80 & 5.02 & 5.51 & 2.25 \\
& \hspace{1mm} + vis. & 2.32 & 2.63 & 1.49 & 2.21 & 2.49& 1.41 & 2.10 & 2.50 & 2.17 \\
& \hspace{2mm} + UBD (\sysname{} global) & \textbf{1.70} & \textbf{1.91} & \textbf{1.03} & \textbf{1.24} & \textbf{1.29} & \textbf{0.38} & \textbf{1.64} & \textbf{1.89} & \textbf{1.16} \\
\bottomrule
\end{tabular}
}
\end{table}

\subsection{Model accuracy}
In this section, the training and test data are from the same datasets (i.e. same capture setups). 
We compare geometry accuracy for different methods on FaMoS (Tab.~\ref{tab:sota_famos} and Fig.~\ref{fig:famos_geometry_accuracy}) and FaceScape (Tab.~\ref{tab:sota_facescape} and Fig.~\ref{fig:fs_geometry_accuracy}) datasets.
As expected, all the multiview face capture methods outperform DECA~\cite{deca}, a 3DMM-based single-image face reconstruction method trained to recover the face region.
Note that DECA results are obtained with Proteus analysis (i.e. rigid alignment) with scale as in~\cite{deca}.
The most accurate prediction from all the views is adopted for metric calculation.
The face region error of DECA (e.g. 3.33 mm mean) is worse than the reported numbers in other papers (about 2 mm) mainly because FaMoS and FaceScape datasets capture large-scale expressions.

The proposed modules are all helpful and finally \sysname{} achieves similar accuracy with TEMPEH on the FaMoS dataset. A similar trend has been observed on the FaceScape dataset.
On the FaceScape dataset, TEMPEH's performance drops a lot since the FaceScape dataset contains scan noise and its registered meshes are not as accurate as the FaMoS dataset.
\sysname{} outperforms TEMPEH and the ablative settings by a large margin, demonstrating the effectiveness of the proposed ``update-by-disagreement'' learning strategy when noise exists.

The scalp of the mesh reconstructed by TEMPEH (global and full) on the FaceScape dataset is flat (see Fig.~\ref{fig:fs_geometry_accuracy}), which is caused by the inaccurate localization result.
``+ V.H. init.'' does not face this problem, which shows the non-data-driven visual hull-based head localization is more robust when the training data are noisy.
``+ V.H. init.'' further improves the estimated expression (note the mouth region).

Since we use per-viewpoint normalized depth (according to the min/max depth of the voxel grid) in Eq.1, setting $\rho=0.1$ results in a margin varying from 3.2 to $3.2\sqrt{3}$ cm, which is large enough to tolerate inaccurate facial details and accurate enough to exclude harmful information from the other side (e.g. image took from left for the right ear).
If the parameter is too large, the method is equivalent to not using the visibility information (a.2 in Tab.\ref{tab:sota_facescape}).
And we have tested smaller $\rho$ (0 and 0.05) in Tab.~\ref{tab:rho}.
The experiment result shows that our method is robust to different hyperparameters.

\subsection{Model generalization}
\label{sec:generalization}
\begin{figure}[t]
\centering
\includegraphics[width=\textwidth]{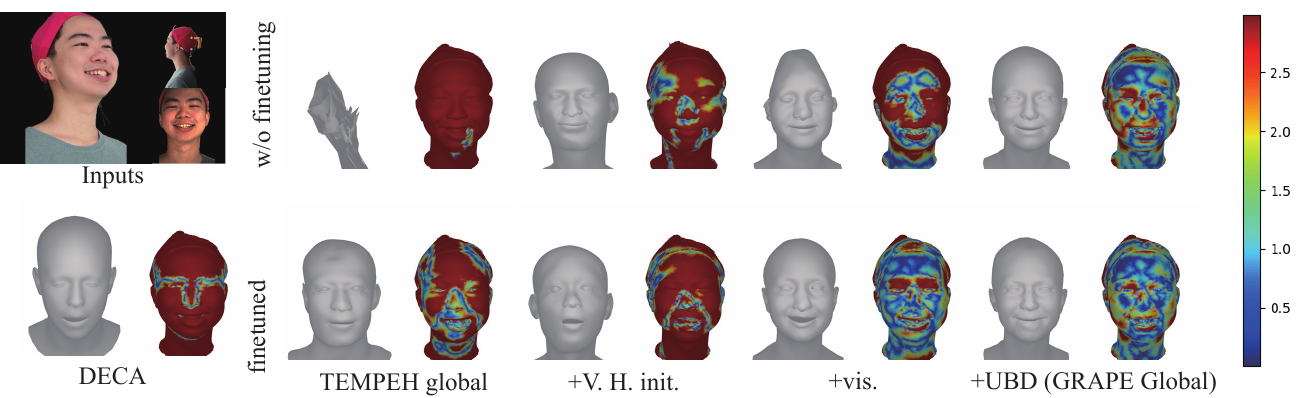}
\caption{\textbf{Qualitative comparison of generalization - FaMoS $\rightarrow$ FaceScape.}
The models are trained on the FaMoS data and evaluated on the FaceScape dataset.
For each model, we show the predicted mesh (left) and the raw scan mesh (right) colored with the point-to-surface distance. The corresponding color bar (3 mm max.) is shown on the right.
}
\label{fig:famos_to_fs}
\end{figure}

In this section, we test the generalization ability of different models in Tab.~\ref{tab:generalization} and Fig.~\ref{fig:famos_to_fs}.
Specifically, we train the networks on one dataset (e.g. FaMoS) and test their accuracies on data captured by a different camera array (e.g. FaceScape).
We report results obtained by direct inference (i.e. w/o finetuning), finetuned on one subject's data (A or B) from the new camera array.
TEMPEH~\cite{tempeh} could not generate a valid result when directly tested on a new dataset (refer to Fig.~\ref{fig:famos_to_fs}).
It still fails to produce a reasonable shape (still obviously worse than DECA) after finetuning its network on one subject's data from the new dataset.
Incorporating our proposed visual hull initialization and visibility-aware 3D feature aggregation gradually improves the generalization of TEMPEH and makes TEMPEH outperform DECA by a substantial margin, demonstrating the remarkable effectiveness of the proposed modules.
As for our \sysname{}, it can be readily used on another dataset.

\begin{figure}[t]
\centering
\includegraphics[width=\textwidth]{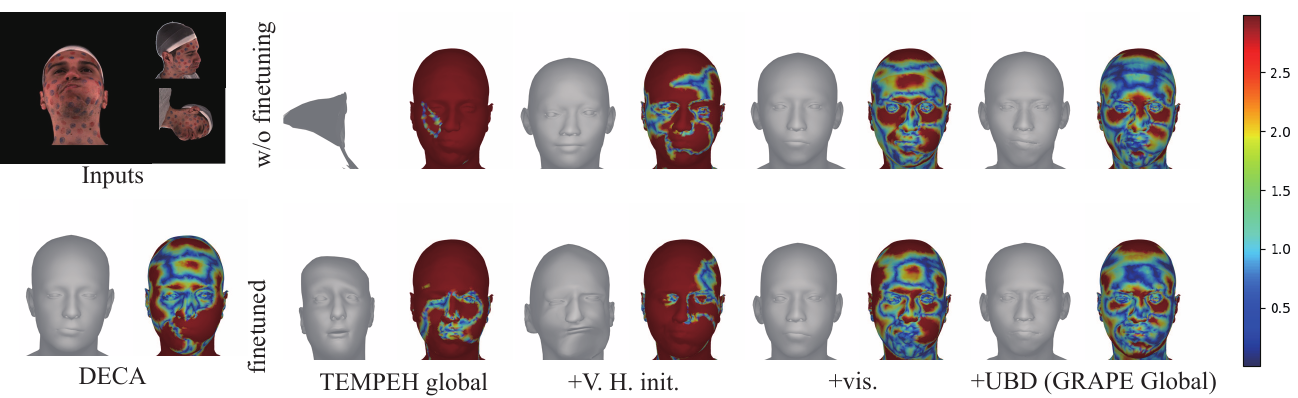}
\caption{\textbf{Qualitative comparison of generalization - FaceScape $\rightarrow$ FaMoS.}
The models are trained on the FaceScape data and evaluated on the FaMoS dataset.
For each model, we show the predicted mesh (left) and the raw scan mesh (right) colored with the point-to-surface distance. The corresponding color bar (3 mm max.) is shown on the right.
}
\label{fig:fs_to_famos}
\end{figure}

\section{Conclusion \& Limitation}
We present a generalizable topologically consistent multi-view head reconstruction method that can be readily applied to new camera arrays and is robust to the noise in the training data, greatly reducing the data collection and processing effort after building a new camera array.
The key to such generalization lies in our generalizable initialization module, which consists of a visual hull based head localizer and a visibility-aware 3D feature aggregation enabled by the visual hull.
To handle commonly seen noise (i.e. imperfect registration, noisy scan) in the training data, we propose an ``update-by-disagreement'' learning strategy.

We focus on generalizing a well-trained network (on sufficient data) to data captured by a different ``camera array''.
However, it still cannot handle other types of out-of-domain (OOD) data as all data-driven methods. 
For example, it still struggles to handle very different unseen expressions missing from the larger training set. 

\clearpage
\section*{Acknowledgement}
This work was supported by the National Natural Science Foundation of China (Grant No.62172126) and the Shenzhen Research Council (Grant No.JCYJ20210
324120202006).
\bibliographystyle{splncs04}
\bibliography{egbib}
\end{document}